\documentclass[10pt,twocolumn,letterpaper]{article}

\usepackage{iccv}
\usepackage{times}
\usepackage{epsfig}
\usepackage{graphicx}
\usepackage{amsmath}
\usepackage{amssymb}
\usepackage{multirow}
\usepackage[inline]{enumitem}
\usepackage{float}

\usepackage[pagebackref=true,breaklinks=true,letterpaper=true,colorlinks,bookmarks=false]{hyperref}

\iccvfinalcopy 

\def\iccvPaperID{3652} 

\ificcvfinal\fi
\begin{document}

\title{Language Features Matter: 
 \protect\\ Effective Language Representations for Vision-Language Tasks}
\author{Andrea Burns\\
\and
Reuben Tan\\
\and
Kate Saenko\\
\and
Stan Sclaroff\\
\and
Bryan A. Plummer\\
\and
Boston University\\
{\tt\small \{aburns4, rxtan, saenko, sclaroff, bplum\}@bu.edu}
}

\maketitle
\begin{abstract}
Shouldn't language and vision features be treated equally in vision-language (VL) tasks? Many VL approaches treat the language component as an afterthought, using simple language models that are either built upon fixed word embeddings trained on text-only data or are learned from scratch. 
We believe that language features deserve more attention, and conduct
experiments which compare different word embeddings, language models, and  embedding augmentation steps on five common VL tasks: image-sentence retrieval, image captioning, visual question answering, phrase grounding, and text-to-clip retrieval. 
Our experiments provide some striking results; an average embedding language model outperforms an LSTM on retrieval-style tasks; state-of-the-art representations such as BERT perform relatively poorly on vision-language tasks. From this comprehensive set of experiments we  propose a set of best practices for incorporating the language component of VL tasks. To further elevate language features, we also show that knowledge in vision-language problems can be transferred across tasks to gain performance with multi-task training. This multi-task training is applied to a new Graph Oriented Vision-Language Embedding (GrOVLE), which we adapt from Word2Vec using WordNet and an original visual-language graph built from Visual Genome, providing a ready-to-use vision-language embedding: \url{http://ai.bu.edu/grovle}.

\end{abstract}
\vspace{-2mm}
\vspace{-3mm}
\section{Introduction}
In recent years many methods have been proposed for vision-language tasks such as image and video captioning~\cite{fang2014captions,krishna2017dense,venugopalan15iccv,vinyalsCVPR2015,xu2015show}, multimodal retrieval~\cite{hendricks17iccv,klein2014fisher,huangArxiv2017,wangTwoBranch2017,Nam_2017_CVPR,vendrov,xuAAAI2019}, phrase grounding~\cite{flickrentitiesijcv,hu2016natural,plummerPLCLC2017,rohrbach2015}, and visual question answering~\cite{fukui16emnlp,VQA,balanced_binary_vqa,VisualMadlibs_bmvc16,VisualMadlibs}.
Language representations for these models tend to be obtained by averaging word embeddings (\eg~\cite{wangTwoBranch2017,plummerPLCLC2017,plummerCITE2017,klein2014fisher}), feeding features representing each word into a LSTM (\eg~\cite{rohrbach2015,xu2015show,xuAAAI2019}), and using word-level or phrase-level attention models (\eg~\cite{andersonCVPR2018, Fang2018ImageCW, Lu:2016:HQC:3157096.3157129, chenEMNLP2018, lee2018stacked}). The word embeddings used in these tasks include a simple one-hot encoding of each word in a vocabulary (\eg~\cite{fukui16emnlp,vinyalsCVPR2015,wangTwoBranch2017}), pretrained dense vector representations like Word2Vec~\cite{w2v} or GloVe~\cite{pennington2014glove}, and Fisher vectors built on top of these dense representations (\eg~\cite{klein2014fisher,plummerCITE2017,wangTwoBranch2017}). Although there are more modern embeddings such as FastText~\cite{bojanowski2017enriching}, ELMo~\cite{Peters:2018} and BERT~\cite{bert} that have shown significant performance improvements on language tasks such as sentiment analysis and question answering, many vision-language approaches still use the more dated feature representations. 

\begin{figure}
    \centering
    \includegraphics[scale=.1]{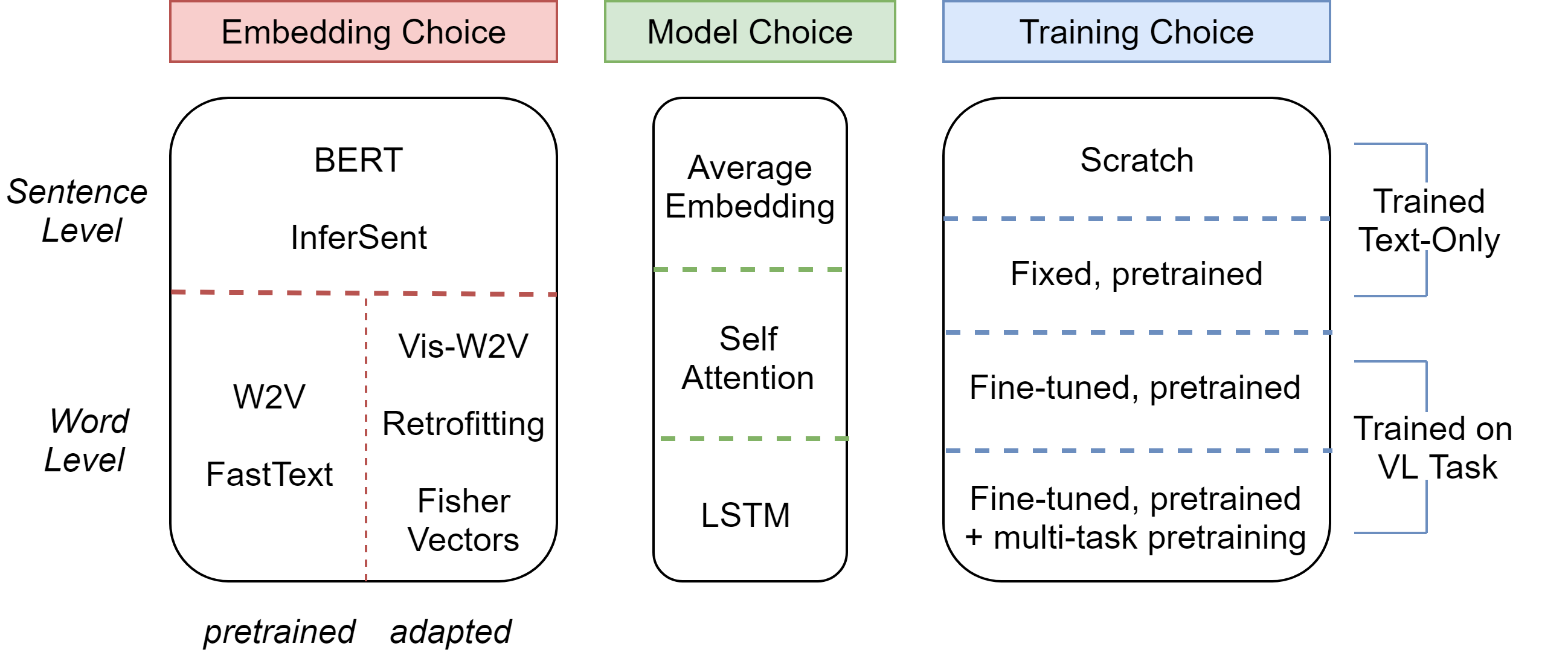}
    \caption{How should language features be constructed for a vision-language task? We provide a side by side comparison of how word-level and sentence-level embeddings, simple and more complex language models, and fine-tuning and post-processing vectors impact performance.}
    \label{fig:motiv}
    \vspace{-4mm}
\end{figure}

While there are isolated cases where these language model and feature choices are compared for the same task model (\eg~\cite{wangTwoBranch2017,hinamiEMNLP2018}), to our knowledge there exists no comprehensive comparison. To address this neglect of language feature exploration, we provide an all-inclusive experimental survey of embedding, language model, and training choice. We perform experiments using from-scratch, Word2Vec~\cite{w2v}, WordNet retrofitted Word2Vec~\cite{FaruquiNAACL}, FastText~\cite{bojanowski2017enriching}, Visual Word2Vec~\cite{visw2v}, HGLMM (300-D, 6K-D)~\cite{klein2014fisher}, InferSent~\cite{infersent}, and BERT~\cite{bert} representations in addition to a new embedding, GrOVLE, on five vision-language tasks: image-sentence retrieval, visual question answering, phrase grounding, image captioning, and text-to-clip retrieval. 

\begin{figure}
    \centering
    \includegraphics[scale=.1]{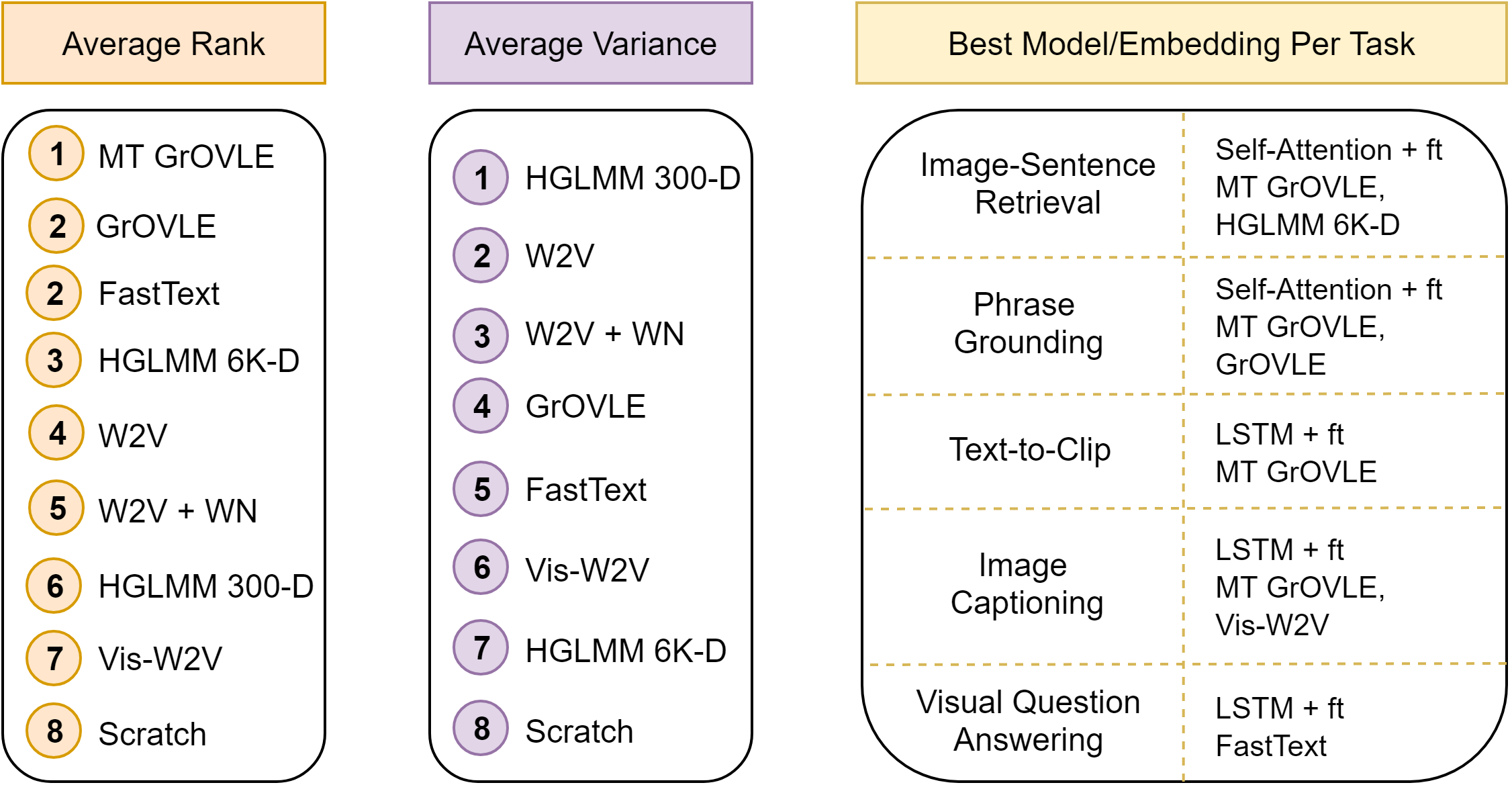}
    \caption{Average rank is defined using each tasks' best performing model. Variance is defined as the average difference between the best and worst performance of the fine-tuned language model options (\eg Average Embedding + ft, Self-Attention + ft, LSTM + ft). Note that variance rank is listed from lowest to highest, \eg from-scratch embeddings have highest variance. If the top embedding per task is a tie, both are provided in the right most column. For the tasks InferSent and BERT operate on, they would land between 7th and 8th place for average rank; average variance is N/A. Note that average variance is not provided for multi-task trained GrOVLE as it was created with the best model for each task.}
    \label{fig:results}
    \vspace{-2mm}
\end{figure}

Our goal is to provide insight for vision-language applications based on extensive experiments varying choices illustrated in Figure \ref{fig:motiv}. Our findings show how to make these choices to take advantage of language features in vision-language work. For example, we find that using an Average Embedding language model, which ignores word ordering, tends to perform better than a LSTM.  This suggests that the LSTM overfits to the task it is trained on. However, when training a word embedding from scratch a LSTM performs best. This result is mostly likely a product of the LSTM learning to predict the next word given previous words, learning context.  Pretrained word vectors likely already provide some semblance of this context information since that is how they are typically trained. The take-aways from all experimental results are summarized in Figure~\ref{fig:results}.



Relying on word embeddings trained solely on large text corpora can have important consequences. For example, in Word2Vec the words ``boy" and ``girl" have higher cosine similarity than either have to the word ``child." While this is a subtle difference, it can impact tasks such as image captioning where ``girl" can be replaced by ``child" when describing a visual scene, but not by ``boy." These nuances are not well captured when using text-only information. To address this, we introduce the Graph Oriented Vision-Language Embedding, GrOVLE, which has been learned for vision-language tasks specifically. 

When building GrOVLE, we take into account the differences in the relationships between words when used to describe visual data. We introduce a new relational graph by extracting semantic relationships between words using the Visual Genome  dataset~\cite{krishnavisualgenome}, which is annotated with dense descriptions of entities, their attributes, and their relationships to other entities within an image. We use both WordNet and Visual Genome graphs to adapt Word2Vec, through the retrofitting process defined by Faruqui~\etal\cite{FaruquiNAACL}.

Finally, in addition to viewing embedding performance for each individual task, we asked: Can an embedding generalize across vision-language tasks? Inspired by multi-task training strategies like PackNet~\cite{packnet}, we train the GrOVLE embedding on all the vision-language tasks in our experiments. The word representation becomes more powerful with task specific knowledge, as the multi-task GrOVLE ultimately outperforms its single-task trained version, becoming a leading embedding amongst the five tasks. Note that unlike PackNet, GrOVLE operates directly on the word embeddings rather than model weights.

\noindent Below we summarize our primary contributions:

\begin{itemize}
    \vspace{-2mm}
    \item Comprehensive experiments exhaustively comparing different word representations, language models, and pretraining and adaptation steps across five common vision-language tasks, providing best practices for future work. See Figure~\ref{fig:results} for a summary of our findings.
    \vspace{-5mm}
    \item GrOVLE, a publicly available word embedding which has been specially trained for vision-language tasks\footnote{\url{http://ai.bu.edu/grovle}}.   
    \vspace{-1mm}
    \item Key insight into the transferability of word embeddings across the five vision-language tasks through the use of multi-task training.
    \vspace{-3mm}
\end{itemize}

\label{subsec: results1}

\section{Related Work}
\label{sec:related}

To the best of our knowledge, the effect of pretrained embeddings in VL tasks has never before been systematically compared. Visual information has been used in limited ways to improve word embeddings such as simply concatenating visual features \cite{Kiela} or focusing on abstract scenes \cite{visw2v}. Lazaridou~\etal\cite{lazaridou} focuses on leveraging first order semantic relationships by encouraging alignment between the visual and language embeddings for a predefined set of nouns describing objects. Word embeddings have also been improved by including additional constraints on the learning process~\cite{YuDredzeAACL} or as a post-processing step~\cite{FaruquiNAACL}. These models focus on improving some general sense of word similarity. GrOVLE is different in that it is directly optimized to work well on a variety of vision-language tasks. We focus on how 10 representations compare amongst model and training choices, some of which are considered state-of-the-art for language tasks such as the recently introduced BERT~\cite{bert}.

Several vision-language approaches have also tried to improve their language model, rather than the word embeddings, as a way to improve performance.  These have included building Fisher vectors on top of pretrained word embeddings~\cite{klein2014fisher,klein2015rnn}, constraining a coarse-to-fine word ordering~\cite{FaghriArxiv2017,vendrov}, or performing co-reference resolution to identify additional constraints between entities (\cite{wang2016matching,plummerPLCLC2017,kong2014you,ChenICCV2017}). Attention mechanisms have also become a popular way to improve performance: word-level attention has been used in image captioning by learning the weights of words using a LSTM~\cite{andersonCVPR2018} or a multi-layered perceptron~\cite{xu2015show,Fang2018ImageCW} before being passed to a language generation model. Dual attention~\cite{Nam_2017_CVPR} has also been used to attend to the question in VQA using feed-forward neural networks. These approaches could be used in conjunction with this work to further improve performance.
 
\vspace{-2mm}
\section{Language Models}
\label{sec:models}
We present three language model options for which we provide experimental results for 8 of 10 different embeddings to determine which language model is best for each task and each embedding (sentence level embeddings cannot be incorporated into some of these architectures). 

\begin{figure}[t]
    \centering
    \includegraphics[width=0.4\textwidth]{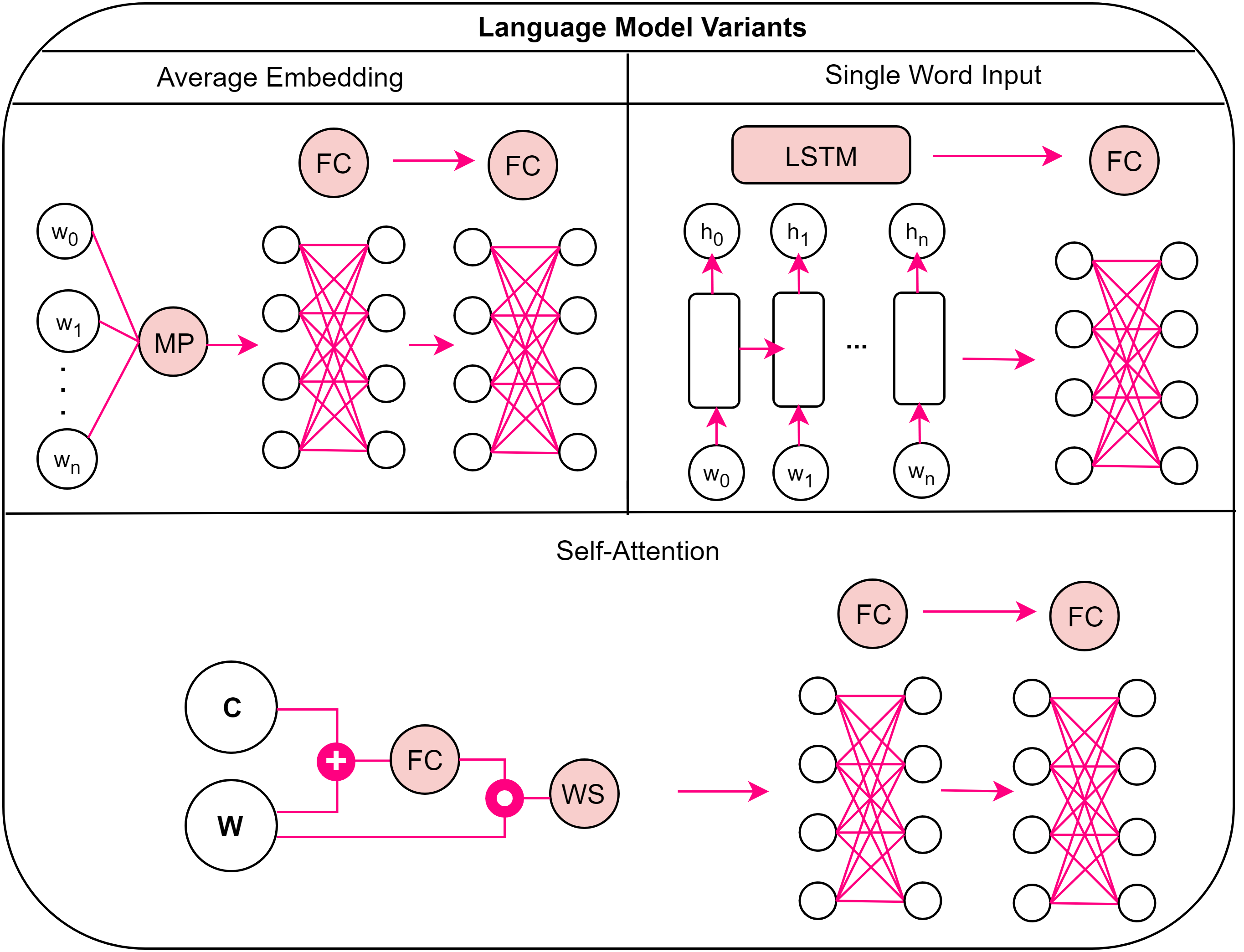}
    \caption{The language model variants used in our experiments include: mean pooling of embeddings (MP) which is then passed to fully connected layers (FC), a LSTM fed a single embedding at a time followed by a fully connected layer, or a self-attention model which builds a weighted context sum (WS) before being passed to a pair of fully connected layers.}
    \label{fig:language_models}
    \vspace{-3mm}
\end{figure}
In Figure~\ref{fig:language_models} an Average Embedding, Self-Attention, and LSTM language architecture are shown. The Average Embedding model consists of mean pooling the embeddings, forming a single representation of all words $w_i$ (with $n$ words in total) in a given sentence or phrase. A sample's pooled vector is then passed through a pair of fully connected layers as shown in the upper left corner of Figure~\ref{fig:language_models}. 

A more complex language architecture is a LSTM; word representations are individually passed through a LSTM cell, each producing their own hidden state. LSTMs are typically thought of as a ``better" architecture choice, modeling the relationship between words in a sentence, as it maintains word ordering. We later show this assumption does not hold true across all vision-language tasks.

Lastly, we compare a Self-Attention model that is closely related to the Average Embedding architecture. The primary difference is the pooling layer, which now consists of two steps. First, a context vector $\textbf{C}$ is concatenated with all word embeddings in $\textbf{W}$ of a given sample. Our experiments use the average embedding as context. It is passed through a fully connected layer which applies Softmax to give context ``scores" for each word in a sentence. Next, the inner product is taken of these weights and the original word embeddings from $\textbf{W}$ to produce a context weighted sum which is then passed to a pair of fully connected layers.

\section{Experimental Setup}
\label{sec:setup}

In this section we provide details of each vision-language task. The datasets and vision-language task models are described in the appendix, but are referenced in Table~\ref{tab:pretrained}.  We split our experiments into three parts: Pretrained Embeddings (Section~\ref{sec:pretrain}), Adapted Embeddings (Section~\ref{sec:adapted}), and Multi-task Trained Embeddings (Section~\ref{sec:specialized}).

\subsection{Compared Tasks and Metrics}
\label{subsec:tasks}

\noindent\textbf{Image-Sentence Retrieval.} The goal is to retrieve relevant sentences given an image, or to retrieve relevant images given a sentence.  It is evaluated using Recall@$K$ where $K = [1, 5, 10]$, resulting in six numbers which measure the performance of the model (three for image-to-sentence and three for sentence-to-image).  We report the average of these six numbers as a measure of overall performance. All six numbers can be found in the appendix.
\smallskip

\vspace{-1mm}
\noindent\textbf{Phrase Grounding.}  In phrase grounding the task is to find the location of a phrase given an image it is known to exist in.  Performance is measured using accuracy, where a box is deemed to be successfully localized if it has at least 0.5 intersection over union (IOU) with the ground truth box.
\smallskip

\noindent\textbf{Text-to-Clip.}  For text-to-clip, the goal is to locate the temporal region (\ie the video clip) that is described by a query.  Performance is measured using a mix of Recall@$K$, where $K = [1, 5]$, and the average IOU the predicted temporal location of a query phrase has with its ground truth temporal segments.  We use the evaluation code provided by Hendricks~\etal~\cite{hendricks17iccv} in our experiments.  We report the average of these three metrics as an overall score; all metrics are reported in the appendix.

\noindent\textbf{Image Captioning.}
The goal of image captioning is to produce natural language which describes an image scene with a well formed sentence. The produced captions are evaluated against a set of reference sentences for each image. We report the commonly used evaluation metric BLEU-4, with CIDEr and METEOR results available in the appendix.  

\noindent\textbf{Visual Question Answering.}
In VQA~\cite{{VQA}}, the goal is to produce a free-form natural language answer given an image and question. This open-ended task consists of three types of questions: yes/no, number and other.  The accuracy of the model is determined by the number of correctly answered questions.  We evaluate on the test-dev set.
\vspace{-3mm}
\smallskip

\smallskip

\section{Pretrained Word Embeddings}
\label{sec:pretrain}
We begin our exhaustive search across language feature choices with pretrained word embeddings. These offer an initial comparison across techniques that do not use forms of post-processing to adapt embeddings, but rather learn vectors with different model architectures and training objectives. Word2Vec, FastText, InferSent, and BERT are reviewed before results are discussed.

\subsection{Word Level Representations}
\noindent\textbf{Word2Vec}~\cite{w2v} is one of the most widespread word embeddings in use since its release. It builds off of the probabilistic feed forward Neural Network Language Model (NNLM) introduced in~\cite{nnlm}, which is composed of input, projection, hidden, and output layers. The input is defined by a 1-out-of-V vector where V is the vocabulary size. The projection matrix is shared amongst all words and the computational complexity between hidden and output layers is reduced using a hierarchical Softmax where the vocabulary is represented as a Huffman binary tree. 

Word2Vec introduced two variations of the NNLM model, with the primary distinction being that the non-linear hidden layer is removed and the projection layer is shared amongst all words, \ie the words are averaged. This leads to the first model, Continuous Bag of Words (CBOW), in which given four previous and four future words, the current word is predicted. The second model, Skip-Gram, instead predicts the context words given the current word. This results in maximizing the classification of a word given the words it is surrounded by. Skip-Gram tends to perform better with a larger range of context words, but this also results in greater computational complexity.
\smallskip

\noindent\textbf{FastText}~\cite{bojanowski2017enriching} is an extension of the Word2Vec model in which the atomic entities of the embeddings are no longer words, but are instead character n-grams. N can be decided given the task and time or space constraints. A word is represented as the sum of its character n-gram vectors in addition to the word vector itself. This change of reference can improve performance due to better representation of rare, misspelled, and out of vocabulary words, as the n-grams create more neighbors for use during training.
\subsection{Sentence Level Representations}
\noindent\textbf{InferSent}~\cite{infersent} uses a bi-directional LSTM with max-pooling to create a sentence-level embedding. It is trained using the Natural Language Inference (NLI) task, in which the goal is to categorize natural language English sentence (premise, hypothesis) pairs into three classes: entailment, contradiction, and neutral. The NLI model architecture separately encodes each sentence of the input pair using a BiLSTM. After, the pair's sentences form a shared representation composed of the concatenation of the vectors, the element-wise product, and the absolute element-wise difference. This vector is then fed into a three-class classifier, defined by several FC layers and a Softmax.
\smallskip

\noindent\textbf{BERT}~\cite{bert} is currently the state-of-the-art word embedding model. Its language encoder is a bi-directional multi-layered Transformer which directly follows the architecture described in~\cite{transformer}. The embedding is trained on two tasks: Masked Language Modeling (MLM) and Next Sentence Prediction. The goal of MLM is to predict the original vocabulary ID of a masked word given its context words. Next Sentence Prediction is the binary classification task of determining if the second sentence is the true next sentence.

\subsection{Results}
\begin{table*}[t]
\centering
\small
\setlength{\tabcolsep}{1.5pt}
\begin{tabular}{|rl|c|c|c|c|c|c|c|c|}
\hline
& Task & \multicolumn{2}{|c|}{Image-Sentence Retrieval} & \multicolumn{2}{|c|}{Phrase Grounding} &  Text-to-Clip & \multicolumn{2}{|c|}{Image Captioning} & VQA\\
\hline
& \multirow{2}{*}{Dataset} & \multirow{2}{*}{Flickr30K~\cite{young2014image}} & \multirow{2}{*}{MSCOCO~\cite{lin2014microsoft}} & Flickr30K & \multirow{2}{*}{ReferIt~\cite{kazemzadeh-EtAl:2014:EMNLP2014}} & \multirow{2}{*}{DiDeMo~\cite{hendricks17iccv}} & \multicolumn{2}{|c|}{\multirow{2}{*}{MSCOCO~\cite{lin2014microsoft}}} & \multirow{2}{*}{VQA~\cite{balanced_vqa_v2}}\\
& &  &  & Entities~\cite{flickrentitiesijcv} &  & & \multicolumn{2}{|c|}{} & \\
\hline
& Method & \multicolumn{2}{|c|}{Embedding Network~\cite{wangTwoBranch2017}} & \multicolumn{3}{|c|}{CITE~\cite{plummerCITE2017}} & \multicolumn{2}{|c|}{ARNet~\cite{tencent}} & EtEMN~\cite{hu2017learning}\\
\hline
& Metric & \multicolumn{2}{|c|}{Mean Recall} & \multicolumn{2}{|c|}{Accuracy} & Average & BLEU-4 & CIDEr & Accuracy\\
\hline
\hline
{\bf (a)} & \textbf{Training from scratch} & & & & & & & & \\
& Average Embedding & 44.3 & 73.7 & 70.46 & 51.70 & 33.02 & -- & -- & --\\
& Self-Attention & 44.6 & 77.6 & 70.68 & 52.39 & 33.48 & -- & -- & --\\
& LSTM & 60.0 & 77.5 & 70.47 & 51.57 & 32.83 & 26.7 & 89.7 & 60.95\\
\hline
{\bf (b)} & \textbf{Word2Vec}~\cite{w2v} & & & & & & & &\\
& Average Embedding & 62.5 & 75.0 & 70.03 & 52.51 & 32.95 &  -- & -- & --\\
& Average Embedding + ft & 71.5 & 78.2 & 70.85 & 53.29 & 32.58 & --  & -- & --\\
& Self-Attention & 63.6 & 75.6 & 70.19 & 52.41 & 33.23 & -- & -- & --\\
& Self-Attention + ft & 71.9 & 79.9 & 70.94 & 53.54 & 33.26 & -- & -- & --\\
& LSTM & 68.5 & 72.5 & 69.83 & 52.86 & 33.73 & \textbf{28.5} & 92.7 & 61.40\\
& LSTM + ft & 69.0 & 78.2 & 70.55 & 53.58 & \textbf{33.94} & \textbf{28.5} & \textbf{94.0} & 61.35\\
  \hline
 {\bf (c)} & \textbf{FastText}~\cite{bojanowski2017enriching} & & & & & & & &\\
 & Average Embedding & 69.2 & 78.5 & 69.75 & 51.27 & 32.45 & -- &--&--\\
 & Average Embedding + ft & 73.0 & \textbf{80.7} & 70.62 & 53.24 & 32.01 & -- & -- & --\\
 & Self-Attention & 69.5 & 78.6 & 69.87 & 52.49 & 33.31 & -- & -- & --\\
 & Self-Attention + ft & \textbf{73.1} & 80.6 & \textbf{71.23} & 53.87 & 33.17 & -- & -- & --\\
 & LSTM & 69.1 & 76.9 & 69.76 & 52.21 & 33.06 & \textbf{28.5} & 92.7 & \textbf{61.86}\\
 & LSTM + ft & 68.5 & 80.1 & 71.09 & \textbf{53.95} & 32.51 & 28.3 & 93.2 & 61.66\\
 \hline
 {\bf (d)} & \textbf{Sentence-Level} & & & & & & & & \\
 & InferSent~\cite{infersent} & 71.2 & 76.4 & 57.83 & 52.29 & 31.87 & -- & -- & --\\
 & BERT~\cite{bert} & 71.8 & 75.4 & 69.38 & 50.37 & 32.46 & -- & -- & --\\
\hline
\end{tabular}
\caption{Word Embedding Comparison Across Vision Language Tasks. \textbf{(a)} contains the results of learning an embedding from scratch \ie random initialization with fine-tuning during training.  The remaining sections compare \textbf{(b)} Word2Vec, \textbf{(c)} FastText, and \textbf{(d)} sentence level embeddings InferSent and BERT. All experiments show three model variants: Average Embedding, Self-Attention, and LSTM, with and without fine-tuning during training. Average Embedding and Self-Attention are not used in generation tasks for Image Captioning and VQA as they are known to show worse performance; sentence level embeddings are not applicable for these tasks. See text for discussion.}
\label{tab:pretrained}
\end{table*}

We start with an embedding learned from scratch with random initialization as our first baseline. Results demonstrate that while many previous works use scratch embeddings, this greatly impacts performance in vision-language tasks. Unsurprisingly, when comparing the first lines of Table~\ref{tab:pretrained}(a,b), we find that using Word2Vec rather than an embedding trained from scratch tends to improve performance.  This is more important when considering a larger vocabulary as seen comparing phrase grounding experiments on DiDeMo and ReferIt, whose embeddings trained from scratch using their smaller vocabulary compare favorably to Word2Vec. 

The original Word2Vec embedding pretrained on Google News can be considered a second baseline. While FastText is a more modern embedding, Word2Vec only falls behind within a point or two across all tasks, and even outperforms or performs equally as well as FastText for certain tasks (\eg text-to-clip, image captioning). This validates works which extend Word2Vec such as Retrofitting, HGLMM Fisher Vectors, and GrOVLE, as Word2Vec may still provide advantages with additional adaptations; results for adapted embeddings follow in Section~\ref{sec:adapted}.


Table~\ref{tab:pretrained} also contains a comparison of language model variants across the five vision-language tasks we evaluate on.  We see that fine-tuning a word embedding on a vision-language task can have dramatic effects on the performance of the language model (\eg 5-10\% increase to mean recall on image-sentence retrieval). 


When comparing the architecture choices from Figure~\ref{fig:language_models} we see that for retrieval-based tasks (\ie where the output is not free-form text) the Average Embedding and Self-Attention models perform better than a simple LSTM-based approach, with Self-Attention being best on average. This is especially notable since these two models have fewer parameters and are faster to compute than a LSTM. Choosing to use a Self-Attention language model in future vision-language work will not only boost metrics, but will also be a more time efficient option. The only apparent exception to this is the text-to-clip task. This may be because it is a video-based task which contains some temporal language in its queries~\cite{hendricks17iccv}, so the ordering of words may be especially important to identifying which video clip to select compared to other retrieval-based tasks. While all language models perform closely on ReferIt phrase grounding, this still suggests that there is no need to use the more complex LSTM language model without additional modification.


Lastly, sentence level embeddings InferSent and BERT are compared in Table~\ref{tab:pretrained}(d); results are without fine-tuning. Fine-tuning would likely improve performance, but is difficult to incorporate due to size (\eg the larger BERT model contains a total of 340M parameters while the well-known VGG-16 network uses 138M; fine-tuning the top layers of BERT still requires loading the full model). The two are comparable to each other with the exception of phrase grounding accuracy on Flickr30K Entities; BERT surprisingly outperforms InferSent by 11.55\%. Both InferSent and BERT do not provide the best results across any task, and thus are not a leading option for vision-language tasks. 

 InferSent and BERT reach comparable values to the best Word2Vec models for image-sentence retrieval on Flickr30K, performing more poorly for the MSCOCO dataset. For the remaining retrieval tasks, metrics are below the best performing model and embedding combination within 1-3 points, again noting the unusual exception of InferSent on phrase grounding of Flickr30K Entities, which significantly drops below scratch performance.

\section{Adapted Word Embeddings}
\label{sec:adapted}

Since the introduction of Word2Vec, several enhancement techniques have been proposed. In this section we explore adaptations of Word2Vec which use different methods to post-process embeddings. Extensions either use language enhancements, visual enhancements, or both (\eg WordNet retrofitting, HGLMM vs. Visual Word2Vec vs. GrOVLE, respectively). We shall now briefly discuss these enhancements.

\subsection{Visual Word2Vec}
Visual Word2Vec~\cite{visw2v} is a neural model designed to ground the original Word2Vec representation with visual semantics. Its goal is to maximize the likelihood of a visual context given the set of words used to describe it, thus pushing word representations used to describe the same visual scene closer together. Clusters are first learned offline using features from abstract clip-art scenes such as the locations of objects, pose, expressions, and gaze to provide surrogate class labels. Word vectors initialized with Word2Vec are then passed through a single hidden layer network. After, a learned output weight matrix and Softmax are applied to predict the visual semantic class the words belong to. 
\subsection{HGLMM Fisher Vectors}
Another post-processed embedding we use for this set of experiments is the Hybrid Gaussian-Laplacian Mixture Model (HGLMM) representation built off of Fisher vectors for Word2Vec~\cite{klein2014fisher}. While bag-of-words pooling is simple and commonly applied, Fisher vectors change this pooling technique and achieve state-of-the-art results on many applications. Fisher vectors instead concatenate the gradients of the log-likelihood of local descriptors (which in this case are the Word2Vec vectors) with respect to the HGLMM parameters. HGLMM is a weighted geometric mean of the Gaussian and Laplacian distributions and is fit using Expectation Maximization. Following~\cite{wangTwoBranch2017,plummerCITE2017}, we reduce the dimensions of the original encodings (18K-D) to 6K-D or 300-D using PCA, as it has been found to improve numerical stability on VL tasks (except for experiments on ReferIt which we reduce to 2K-D due to its small vocabulary size).  

\subsection{GrOVLE: Graph Oriented Vision-Language Embedding}
We provide a new embedding, GrOVLE, which adapts Word2Vec using two knowledge bases: WordNet and Visual Genome. This builds off of the retrofitting work of~\cite{FaruquiNAACL} in which WordNet was one of the lexicon options. The Visual Genome relational graph is novel, as it creates a language graph that captures how words are used in visual contexts, unlike any of the language databases used in~\cite{FaruquiNAACL}. We briefly review retrofitting and then detail the construction of our original Visual Genome word relation graph. GrOVLE provides a vision-language enhanced embedding and outperforms Visual Word2Vec across many tasks. The released version of GrOVLE is multi-task trained, creating an additional level of VL knowledge, later described in Section~\ref{sec:specialized}.
\vspace{-6mm}
\label{sec:retro}
\subsubsection{Retrofitting Word Embeddings}
\label{subsec:retrodetails}
In this section we review the approach of Faruqui~\etal\cite{FaruquiNAACL}, which proposed a graph based learning technique to ``retrofit" additional semantic knowledge onto pretrained word embeddings. 

Given a vocabulary $V$ with words $\{w_1, w_2, ..., w_n\}$ and its corresponding word embedding $\hat{Q}$, where $\hat{q}_i$ is the embedding for $w_i$, belief propagation is performed to obtain a new embedding $Q$ which minimizes the distances between the embedding representing each word and its neighbors. These neighbors are defined as edges $E$ between words in a graph.  $L2$ regularization is performed between the original and new word embeddings to help prevent overfitting. We find that this $L2$ regularization is necessary whenever we are updating the word embeddings (\ie we also use it during multi-task training described in Section~\ref{sec:specialized}). We use the same regularization parameters as Faruqui~\etal and refer the reader to their work to view the final objective function.

\begin{table*}[t]
\centering
\small
\setlength{\tabcolsep}{3pt}
\begin{tabular}{|rl|c|c|c|c|c|c|c|c|c|c|c|c|c|c|c|}
\hline
& Task & \multicolumn{2}{|c|}{Image-Sentence Retrieval} & \multicolumn{2}{|c|}{Phrase Grounding} &  Text-to-Clip & \multicolumn{2}{|c|}{Image Captioning} & VQA\\
\hline
& \multirow{2}{*}{Dataset} & \multirow{2}{*}{Flickr30K} & \multirow{2}{*}{MSCOCO} & Flickr30K & \multirow{2}{*}{ReferIt} & \multirow{2}{*}{DiDeMo} & \multicolumn{2}{|c|}{\multirow{2}{*}{MSCOCO}} & \multirow{2}{*}{VQA}\\
& &  &  & Entities &  &  &  \multicolumn{2}{|c|}{} & \\
\hline
& Metric & \multicolumn{2}{|c|}{Mean Recall} & \multicolumn{2}{|c|}{Accuracy} & Average & BLEU-4 & CIDEr & Accuracy\\
\hline
\hline
{\bf (a)} & \textbf{Word2Vec + wn}~\cite{FaruquiNAACL} & & & & & & & &  \\
& Average Embedding + ft & 72.0 & 79.2 & 70.51 & 53.93 & 33.24 & -- & -- & --\\
& Self-Attention + ft & 72.4 & 80.0 & 70.70 & 53.81 & 33.65 & -- & -- & -- \\
& LSTM + ft & 69.3 & 78.9 & 70.80 & 53.67 & 34.16 & 28.6 & 93.3 & 61.06\\
\hline
{\bf (b)} & \textbf{GrOVLE} & & & & & & & &\\
& Average Embedding + ft & 72.3 & 80.2 & 70.77 & \textbf{53.99} & 33.71 & -- & -- & --\\
& Self-Attention + ft & 72.1 & 80.5 & 70.95 & 53.75 & 33.14 & -- & -- & --\\
& LSTM + ft & 69.7 & 78.8 & 70.18 & \textbf{53.99} & 34.47 & 28.3 & 92.5 & 61.22\\
\hline
{\bf (c)} & \textbf{Visual Word2Vec}~\cite{visw2v} & & & & & & & & \\
& Average Embedding + ft & 66.8 & 78.7 & 70.61 & 53.14 & 31.73 & -- & -- & --\\
& Self-Attention + ft & 68.8 & 79.2 &\textbf{71.07} & 53.26 & 31.15 & -- & -- & --\\
& LSTM + ft & 66.7 & 74.5 & 70.70 & 53.19 & 32.29 & \textbf{28.8}& \textbf{94.0} & 61.15\\
 \hline
{\bf (d)} & \textbf{HGLMM (300-D)}~\cite{klein2014fisher} & & & & & & & & \\
& Average Embedding + ft & 71.0 & 79.8 & 70.64 & 53.71 & 32.62 & -- & -- & --\\
& Self-Attention + ft & 71.8 & 80.4 & 70.51 & 53.83 & 33.44 & -- & -- & --\\
 & LSTM + ft & 69.5 & 77.9 & 70.37 & 53.10 & 33.85 & 28.7 & \textbf{94.0} & \textbf{61.44}\\
  \hline
{\bf (e)} & \textbf{HGLMM (6K-D)}~\cite{klein2014fisher} & & & & & & & & \\
 & Average Embedding + ft & 73.5 & \textbf{80.9} & 70.83 & 53.36 & 32.66 & -- & -- & --\\
 & Self-Attention + ft & \textbf{75.1} & 80.6 & 71.02 & 53.43 & 33.57 & -- & -- & --\\
 & LSTM + ft & 68.0 & 79.4 & 70.38 & 53.89 & \textbf{34.62} & 28.0 & 92.8 & 60.58\\
\hline
\end{tabular}
\caption{Modifications of Word2Vec. \textbf{(a)} contains Word2Vec retrofitted results using only the WordNet (wn) lexicon from~\cite{FaruquiNAACL}. Next, \textbf{(b)} is our baseline embedding which includes the new Visual Genome relational graph. Visual Word2Vec results are provided in \textbf{(c)}, and \textbf{(d)}, \textbf{(e)} are Fisher vectors on top of Word2Vec. See text for discussion.
}
\vspace{-1mm}
\label{tab:adapted}
\end{table*}

\vspace{-3mm}
\subsubsection{Word Relation Graph Construction}
\label{subsubsec:graph}
Below we describe the methods we use to create the edges between words which share some semantic relation. We use these edges to retrofit the word embeddings with the process described in Section~\ref{subsec:retrodetails}. Of the lexicons provided by Faruqui~\etal\cite{FaruquiNAACL}, we used only the WordNet graph, as it contains the largest vocabulary with the most edges. A joint lexicon is built with WordNet and Visual Genome as opposed to successively retrofitting the two; this minimized forgetting of the first and thus improved performance.

\smallskip

\noindent\textbf{WordNet}~\cite{wordnet} is a hierarchical lexical database which organizes nouns, adjectives, verbs and adverbs into sets of synonyms (\textit{synsets}) and uses semantic relations to associate them. As in Faruqui~\etal\cite{FaruquiNAACL}, we construct a graph by creating links between words if they have a synonym, hypernym, or hyponym relationship.
\smallskip

\noindent\textbf{Visual Genome} 
\label{subsubsec:visgenome}~\cite{krishnavisualgenome} contains a wealth of language annotations for 108K images: descriptions of entities in an image, their attributes, relationships between multiple entities, and whole image and region-based QA pairs. Each instance in these annotations is considered a sample which we tokenize and remove stopwords from. We compute co-occurrence statistics over pairs of words within the sample for pairs that occur more than 50 times, resulting in 322,928 pairs for 12,849 words. For each word we compute a pointwise mutual information (PMI) score for all pairs it occurs in, and create links between the top ten words. This creates a graph where words that occur frequently together when describing visual data are linked.

 \begin{table*}[t]
\centering
\small
\setlength{\tabcolsep}{3pt}
\begin{tabular}{|rl|c|c|c|c|c|c|c|c|c|c|c|c|c|c|c|}
\hline
& Task & \multicolumn{2}{|c|}{Image-Sentence Retrieval} & \multicolumn{2}{|c|}{Phrase Grounding} &  Text-to-Clip & \multicolumn{2}{|c|}{Image Captioning} & VQA\\
\hline
& Metric & \multicolumn{2}{|c|}{Mean Recall} & \multicolumn{2}{|c|}{Accuracy} & Average & BLEU-4 & CIDEr & Accuracy\\
\hline
& GrOVLE w/o multi-task pretraining & 64.7 & 75.0 & 70.53 & 52.15 & 34.45 & 28.5 & 92.7 & 61.46\\
& + multi-task pretraining w/o target task & 65.8 & 76.4 & 70.82 & 52.21 & 34.57 & \textbf{28.8} & \textbf{93.3} & 61.47\\
& + multi-task pretraining w/ target task & 66.2 & 80.2 & 70.87 & 52.64 & 34.82 & 28.5 & 92.7 & \textbf{61.53}\\
& + multi-task pretraining w/ target task + ft &  \textbf{72.6} &  \textbf{81.3} & \textbf{71.57} &  \textbf{54.51} &  \textbf{35.09} & 28.7 & 93.2 & 61.46\\
\hline
\end{tabular}
\caption{Comparison of training our word embeddings on four tasks and testing on the fifth, as well as training on all five tasks.}
\label{tab:multi_task}
\vspace{-2mm}
\end{table*}

\begin{table*}
\centering
\small
\setlength{\tabcolsep}{3pt}
\begin{tabular}{|rl|c|c|c|c|c|c|c|c|c|c|c|c|c|c|c|}
\hline
& Task & \multicolumn{2}{|c|}{Image-Sentence Retrieval} & \multicolumn{2}{|c|}{Phrase Grounding} &  Text-to-Clip & \multicolumn{2}{|c|}{Image Captioning} & VQA\\
\hline
& Additional Models & \multicolumn{2}{|c|}{SCAN  \cite{lee2018stacked}} &  \multicolumn{2}{|c|}{QA R-CNN \cite{hinamiEMNLP2018}} & TGN \cite{chenEMNLP2018} & \multicolumn{2}{|c|}{BUTD \cite{andersonCVPR2018}} & BAN\cite{kim2018bilinear}\\
\hline
& Metric & \multicolumn{2}{|c|}{Mean Recall} & \multicolumn{2}{|c|}{Accuracy} & Average & BLEU-4 & CIDEr & Accuracy\\
\hline
& Training from scratch & 72.8 & 83.2 & 68.56 & 50.23 & 43.91 & 35.2 &109.8 & 68.98 \\
& FastText + ft & 72.5 & 83.8 & 69.27 & 53.01 & 44.21 & 35.2 & 110.3& 69.91 \\
& GrOVLE (w/o multi-task pretraining) + ft & 72.7 & 84.1 & 70.03 & 53.88 & \textbf{45.26} & 35.1 & 110.4& 69.36\\
& + multi-task pretraining w/ target task + ft & \textbf{76.2} & \textbf{84.7} & \textbf{71.08} & \textbf{54.10} & 43.61 & \textbf{35.7} & \textbf{111.6} & \textbf{69.97} \\
\hline
\end{tabular}
\caption{We include results with additional models to verify trends. See text for discussion and the appendix for more.}
\label{tab:moremodels}
\vspace{-3mm}
\end{table*}
\vspace{-1mm}
\subsection{Results}
\label{subsec:results2}
We see a small, but consistent improvement across most of the vision-language tasks using GrOVLE as seen in Table~\ref{tab:adapted}(b). These changes result in an embedding with comparable performance to the HGLMM 6K-D features, which are reported in Table~\ref{tab:adapted}(e).   However, our word embedding tends to perform better when embeddings are the same size (\ie 300-D). For the generation-based tasks (\ie captioning and VQA), the benefits of using adapted embeddings are less clear.  This may simply be an artifact of the challenges in evaluating these tasks (\ie, the captions are improving in a way the metrics don't capture).  Also, models that more carefully consider the effect of each word in a caption may benefit more from our improved features (\eg~\cite{Nam_2017_CVPR,xuAAAI2019}).

While Visual Word2Vec is an established visually-enhanced embedding, its published results did not include these vision-language tasks. Visual Word2Vec performs comparably amongst results for generation tasks (\ie image captioning and VQA), but these tasks have little variance in results, with less than a point of difference across the adapted embeddings. The small gain provided in generation tasks by Visual Word2Vec does not out-weight the drops in performance across other tasks such as the significant mean recall drop of 6.3 compared to HGLMM's 6K-D Self-Attention result in line two of Table~\ref{tab:adapted}(c) and Table~\ref{tab:adapted}(e) for image-sentence retrieval of Flickr30K. For comparison, GrOVLE's Self-Attention result in Table~\ref{tab:adapted}(b) is only 3 points lower.

Finally, we report results using HGLMM of different dimension. HGLMM 300-D features are used for a more fair comparison to other embeddings. While the HGLMM 6K-D representation primarily results in the highest performance, it performs more poorly on generation tasks and also results in high variance. For example, column one in Table~\ref{tab:adapted}(e) shows a range of 7.1 in mean recall, unlike GrOVLE which has a range of 2.6. 
\vspace{-2mm}
\section{Multi-task Training}
\label{sec:specialized}

A drawback of using pretrained word embeddings like Word2Vec or the retrofitting process is that they are trained solely on text data.  While our Visual Genome Graph provides some general information on how words in our vocabulary are used for visual data, it doesn't provide any sense of visual similarity between semantically different words that may be necessary to perform a particular vision-language task.  To address this, we fine-tune GrOVLE across the five VL tasks. 

We provide results for a four and five multi-task trained embedding. The four task experiments are performed with the final task embedding fixed to demonstrate how well the embeddings would generalize to new tasks. We also provide results for pretraining on five tasks with and without fine-tuning during the last task. Similarly to PackNet~\cite{packnet}, for each dataset/task in the four and five task experiments, we keep the $K$ most informative features frozen when training any subsequent task, diminishing the effect of catastrophic forgetting when fine-tuning on a new task. For an embedding of size $D$ and $T$ tasks, $K=\frac{D}{T}$, \ie $K = 60$ in our experiments. We evenly split the $K$ features for tasks with multiple datasets.  Features that were tuned on a task are ranked according to variance and frozen before training on the next dataset/task.  The end result is a pretrained word embedding which can be ``dropped in'' to existing models to improve performance across many vision-language tasks.

To verify that the multi-task GrOVLE performance improvements generalize across task model architecture, we provide results using additional task models in Table~\ref{tab:moremodels}. More results can be found in the appendix.

\subsection{Results}

\label{subsec:results3}
Table~\ref{tab:multi_task} reports results of the multi-task training procedure described above. We use the best performing language model in our comparisons for each task, \ie Self-Attention for image-sentence retrieval and phrase grounding, and the LSTM language model for text-to-clip, image captioning, and VQA. The first lines of Table~\ref{tab:multi_task} report the results of the original fixed GrOVLE embedding, which should be considered the baseline. The second line of Table~\ref{tab:multi_task} reports performance when the four-task pretrained GrOVLE is fixed when used in the target task, \ie the task currently being run. The third and fourth line of Table~\ref{tab:multi_task} report the results of our embedding when they were trained on all five tasks, and kept fixed or fine-tuned for the target task, respectively. 

The results of line three and four demonstrate that our improved embedding tends to transfer better when applied with fine-tuning during the target task. We find similar trends in performance improvements across tasks: larger gains occur for image-sentence retrieval with +7.9 mean recall for the Flickr30K dataset and +6.3 for MSCOCO. All other tasks have performance improvements under one point, showing that while the vision-language tasks appear to transfer well without harming performance, they are leveraged most in image-sentence retrieval, with an exception of phrase grounding accuracy on ReferIt (+2.36\%). 

Table \ref{tab:moremodels} provides more models per task and demonstrates consistent results:   embeddings  can  significantly  affect  performance and GrOVLE variants are still the best embedding overall. As we move down the table we find even larger performance improvements made by using the five-task pretrained GrOVLE with fine-tuning than in Table~\ref{tab:multi_task}. This multi-task variant is the best performing across all tasks, thus we release this embedding for public use. 

\vspace{-2mm}



\section{Conclusion}
 We believe there are five major findings in our experiments that researchers should keep in mind when considering the language component for vision-language tasks: 
 \begin{enumerate}
    \item On retrieval-style tasks, the Average Embedding and Self-Attention language model tend to outperform a simple LSTM.
    \item Fine-tuning a word embedding for a task can significantly impact performance.
    \item For standard vision-language metrics, language features matter most on retrieval and grounding tasks, and less on text-to-clip and generation tasks.
    \item Word embeddings trained on outside vision-language datasets and tasks generalize to other applications.
    \item Multi-task trained GrOVLE is the leading embedding option for four of the five vision-language tasks when used with the best corresponding language model.
\end{enumerate}

We have provided evidence that language and vision features should be treated equally when used in vision-language tasks. When using the best embedding, language model, and training choices, performance for tasks with more variance can greatly improve, and tasks with more stubborn performance metrics can be nudged further. These insights are proposed to benefit future vision-language work. Along with these findings, we have introduced GrOVLE, which incorporates hierarchical language relations from WordNet as well as language with visual context from Visual Genome. In addition to these adaptations, we perform multi-task training with five common vision-language tasks to further incorporate nuanced visual information. This provides a 300-D embedding with vision-language enhancements that is comparable to current embeddings and provides low variance results. 


\section*{Acknowledgements}
We would like to thank the reviewers for their helpful suggestions. This work is supported in part by DARPA and NSF awards IIS-1724237, CNS-1629700, CCF-1723379.
\vspace{-2mm}
{\small
\bibliographystyle{ieee_fullname}
\bibliography{ms}
}
\onecolumn
\section{Appendix}
\subsection{Datasets}
\label{subsec:datasets}

\noindent\textbf{Flickr30K~\cite{young2014image}.}  This dataset consists of 32K images obtained from the Flickr website, each of which has been annotated with five descriptive captions. We use the splits of Plummer~\etal~\cite{flickrentitiesijcv}, which separate the dataset into 30K/1K/1K train/test/validation images which we use for the image-sentence retrieval and phrase grounding tasks.
\smallskip

\noindent\textbf{MSCOCO~\cite{lin2014microsoft}.}  This dataset links 123K images for the training and validation sets (80K/40K images, respectively), each of which is annotated with five descriptive captions.  For the image-sentence retrieval experiments, we use the test/validation splits from Wang~\etal~\cite{wangTwoBranch2017}, which consists of 1K images for each split, for a total of 2K images, randomly sampled from the validation set. For image captioning experiments, use the splits from Chen~\etal~\cite{tencent}, which reserves 5K images each for validation and testing.
\smallskip

\noindent\textbf{Flickr30K Entities~\cite{flickrentitiesijcv}.}  This dataset augments the Flickr30K dataset with 276K bounding boxes which are linked to noun phrases in the descriptive captions.  We use the same splits as the Flickr30K dataset, resulting in 14.5K instances across the 1K images in the test set for the phrase grounding task.  Following~\cite{flickrentitiesijcv,rohrbach2015,wangTwoBranch2017}, we use the union of the bounding boxes for the ground truth box of a phrase which is linked to multiple boxes.
\smallskip

\noindent\textbf{ReferIt~\cite{kazemzadeh-EtAl:2014:EMNLP2014}.}  This dataset augments the 20K images from the IAPR RC-12 dataset~\cite{Grubinger06theiapr} with 120K region descriptions. We split the splits of Hu~\etal~\cite{hu2016natural}, which split the images evenly into train/validation and test sets (10K each), resulting in about 60K instances in each split. 
\smallskip

\noindent\textbf{DiDeMo~\cite{hendricks17iccv}.}  This dataset consists of just over 10,000 videos, each of which has between 3-5 video segment descriptions.  We use the splits provided by Hendricks~\etal~\cite{hendricks17iccv}, which splits the videos into sets of 8.4K/1K/1K for train/test/validation.
\smallskip

\noindent\textbf{VQA v2~\cite{balanced_vqa_v2}.}   This dataset augments images from MSCOCO with QA pairs.  The training, validation and test image sets contain 83K, 41K, and 81K images, respectively.  This constitutes 444K, 214K, and 448K questions for training/validation/testing splits.  Each training and validation question has ten answers provided.
\subsection{Task Methods}
\label{subsec:task_methods}

\noindent\textbf{Image-Sentence Retrieval.}  We use a modified implementation of the Embedding Network~\cite{wangTwoBranch2017} provided by the authors in our experiments\footnote{\url{https://github.com/lwwang/Two_branch_network}}.  This model uses two branches, one for text and one for images, to learn a projection to a shared embedding space where Euclidean distance is used to measure similarity between images and sentences.  We use the default parameters and data processing in the author's implementation, except that we compute the visual representation for each image  using a 152-layer ResNet~\cite{He2015} which has been trained on ImageNet~\cite{imagenet_cvpr09}.  Additionally, we use 448x448 crops rather than the 224x224 pixel crops used by Wang~\etal~\cite{wangTwoBranch2017} as done in prior work, \eg~\cite{Zhang_2018_ECCV,Nam_2017_CVPR}.  Following~\cite{wangTwoBranch2017,Zhang_2018_ECCV,Nam_2017_CVPR}, we keep the CNN parameters fixed for a fair comparison.  By default this model uses an Average Embedding language model.  When we use the LSTM language model, we use a hidden state of 512-D.  We set regularization coefficient $\alpha$ to be 1e-4 when fine-tuning the Average Embedding and Self-Attention model and 1e-6 for the LSTM model.
\smallskip

\noindent\textbf{Phrase Grounding.} To evaluate our word embeddings on this task, we use the implementation of CITE network~\cite{plummerCITE2017}\footnote{\url{https://github.com/BryanPlummer/cite}}.  This model learns a set of embeddings which share some parameters, each of which captures a different concept important for phrase grounding.  Following Plummer~\etal~\cite{plummerArxiv2018}, we use the parameters and feature representation learned from fine-tuning a 101-layer ResNet and Region Proposal Network.  This model also uses an Average Embedding language model by default, and we use 256-D hidden state for our LSTM experiments.  We set regularization coefficient $\alpha$ to be 1e-5 for both datasets.
\smallskip

\noindent\textbf{Text-to-Clip.} When we performed our experiments none of the methods on the DiDeMo dataset which outperform the baseline model of Hendricks~\etal\cite{hendricks17iccv} had publicly available code for the text-to-clip task (\eg~\cite{chenEMNLP2018,Liu_2018_ECCV}).  As a result, we used the CITE network for the text-to-clip task since it performed better than the baseline model as well as better than the phrase-region grounding Similarity Network~\cite{wangTwoBranch2017} and straightforward adaptations of the R-C3D model~\cite{Xu2017iccv} in our experiments.  We learn $K=8$ concept embeddings for this dataset and use the VGG~\cite{simonyan2014very} features for the visual representation provided by Hendricks~\etal\cite{hendricks17iccv}.  We use a 512-D hidden state for our LSTM models, and set regularization coefficient $\alpha$ to 5e-2.  This dataset likely required additional regularization when fine-tuning its embeddings due to its relatively small size.
\smallskip

\noindent\textbf{Image Captioning.}
We use a PyTorch implementation
\footnote{\url{https://github.com/chenxinpeng/ARNet}}
of the Auto-Reconstructor Network (ARNet) architecture~\cite{tencent} provided by the authors. This model builds off of the original Neural Image Captioning (NIC) architecture~\cite{vinyalsCVPR2015} by adding an additional LSTM to reconstruct previous hidden states. We set the regularization coefficient of the NIC loss, $\alpha$, to be 5e-2 when fine-tuning the word embeddings. ARNet's additional stacked LSTM takes a current hidden state as input and attempts to generate the previous hidden state. This can be viewed as a ``soft" zoneout strategy as the model adaptively learns how to reconstruct the last hidden state at each time step, as opposed to the typical zoneout regularizer which makes a binary choice between previous and current hidden states.
\smallskip

\noindent\textbf{Visual Question Answering.}
We use the authors' implementation\footnote{\url{https://github.com/ronghanghu/n2nmn}} of the End-to-End Module Networks~\cite{hu2017learning} as our VQA model. This network learns to decompose natural language questions into sub-tasks and assembles question-specific deep networks from neural modules to solve its corresponding sub-task. 
The training process of this model consists of two parts: the cloning expert and the policy search. Since the policy search improves the model by only 0.7\% while adding significant training time, we report results only using the cloning expert.  We use the default parameters in the implementation and follow the authors' data pre-processing steps. When we include L2 regularization on the word embeddings, we set its weight to be 5e-4. Note that we report results using the VQA v2 dataset, whereas  Hu~\etal~\cite{hu2017learning} reported results on VQA v1.

\subsection{Additional Task Methods}
\noindent\textbf{Image-Sentence Retrieval.} We also report results with the Stacked Cross Attention Network (SCAN) model~\cite{lee2018stacked} using the authors' provided implementation\footnote{\url{https://github.com/kuanghuei/SCAN}}.  Unlike the Embedding Network, this model uses the top 36 region-level features~\cite{andersonCVPR2018} which have been trained to capture image concepts on the Visual Genome dataset~\cite{krishnavisualgenome}.  A similarity score is computed between all combinations of words in a sentence and image regions, and then aggregated using a multi-step attention mechanism to obtain an overall matching score.  For each dataset, we use the settings for the best performing single model reported in their paper, \ie, \emph{i-t AVG ($\lambda$1 = 4)} for Flickr30K and \emph{t-i AVG ($\lambda$1 = 9)} for MSCOCO.
\smallskip

\noindent\textbf{Phrase Grounding.} To supplement our results, we experiment with using the implementation of the Query Adaptive R-CNN network~\cite{hinamiEMNLP2018} from Plummer~\etal~\cite{plummerArxiv2018}.  This model adapts Faster R-CNN~\cite{renNIPS15fasterrcnn} to the phrase grounding task.  The implementation in Plummer~\etal updates the VGG network used in the original paper with a 101-layer ResNet, but does not pretrain their model on Visual Genome or use the online hard negative mining~\cite{shrivastavaCVPR16ohem} as done in the original paper.  In addition, Plummer~\etal also reported better performance by randomly sampling 5 phrases associated with an image for each minibatch rather than using all annotated phrases.  We compared this implementation using a VGG network to the grounding performance reported in~\cite{hinamiEMNLP2018} and found it performed similarly on Flickr30K Entities despite these changes, but using a ResNet backbone as done in our experiments does boost performance by 3-8\%.
\smallskip

\noindent\textbf{Text-to-Clip.} We provide additional results from the Temporally Grounding Natural Sentence in Video (TGN) \cite{chenEMNLP2018} model. The TGN model consists of 3 components: the encoder, the interactor and the grounder. Visual and language features are first projected into the same embedding space using the encoder. Next, the interactor computes the frame-by-word interactions using the encoded visual and language features. Finally, based on these interactions, the grounder scores and ranks the temporal segment candidates ending at each frame. We note that these results are obtained from our own implementation of the TGN model as the authors have not released code. In our implementation, we adopt the same hyperparameter values as detailed in \cite{chenEMNLP2018}. 
\smallskip

\noindent\textbf{Image Captioning.} We provide results for two additional image captioning models: the vanilla show-and-tell Neural Image Captioning model (NIC) of Vinyals \etal \cite{vinyalsCVPR2015} and the popular Bottom-Up Town-Down (BUTD) model from Anderson \etal \cite{andersonCVPR2018}.  We set $\alpha$ = 5e-2 as our L2 regularization coefficient when fine-tuning the word embeddings for both models. We use a PyTorch implementation \footnote{\url{https://github.com/yunjey/pytorch-tutorial}} of the NIC model for this task. This model follows an encoder-decoder paradigm inspired by machine translation, in which the probability of a sentence given an image is maximized. A CNN encodes an image which is then fed into a decoder LSTM to form a natural language sentence. Unlike the results reported in Vinyals \etal , we use a single model rather than an ensemble, and use a 152-layer ResNet pretrained on ImageNet as our image encoder.
\smallskip

\noindent We also use a PyTorch implementation \footnote{\url{https://github.com/ruotianluo/self-critical.pytorch}} of the Bottom-Up Top-Down Attention image captioning model. BUTD uses a combination of visual attention mechanisms: bottom-up attention is implemented using Faster R-CNN~\cite{renNIPS15fasterrcnn} to generate object region proposals and their respective features, which are then weighted by the top-down attention mechanism. The model also adds an attribute predictor to Faster R-CNN. The language model is implemented with two standard LSTMs, where the first layer serves as top-down attention and the second is the language generator. The attention LSTM takes the previous time step output, mean pooled image features, and previously generated word encoding as input. After a Softmax is applied to the output of the attention LSTM, the weighted visual features are passed to the generator LSTM.
\smallskip

\noindent\textbf{Visual Question Answering.} We provide additional VQA results using the Bilinear Attention Networks (BAN) model \cite{kim2018bilinear}. The BAN model utilizes adaptive region-level features \cite{andersonCVPR2018} as the visual input. It extracts joint representations from each pair of visual and word features via low-rank bilinear pooling while computing their bilinear interactions using attention maps. We use the provided implementation \footnote{\url{https://github.com/jnhwkim/ban-vqa}} in our experiments and adopt the same hyperparameter settings as described in~\cite{kim2018bilinear}.

\subsection{Discrepancies with Published Work}
If available, we use the authors' publicly available code. Baseline results differ from published values despite this. The best results in~\cite{lee2018stacked},~\cite{vinyalsCVPR2015} are obtained using ensemble methods, but our results use a single model. Although, single model~\cite{lee2018stacked} with the five-task multi-task trained GrOVLE + ft is on par with ensemble results.

\subsection{Comparison of Word2Vec and GloVe}

When initially deciding the set of embeddings to use in our experiments, we did consider GloVe. However, there were insignificant differences between Word2Vec and GloVe results (some shown below). Thus, we didn't include it in the main paper due to space constraints as GloVe is also a dated embedding.

\begin{table}[H]
\centering
\small
\setlength{\tabcolsep}{1.5pt}
\begin{tabular}{|rl|c|c|c|c|}
\hline
& & \multicolumn{2}{|c|}{Image-Sentence Retrieval~\cite{wangTwoBranch2017}} & \multicolumn{2}{|c|}{Phrase Grounding~\cite{plummerCITE2017}}\\
\cline{3-6}
& & Flickr30k & MSCOCO & Flickr30k Entities & ReferIt\\
\cline{3-6}
& Method & \multicolumn{2}{|c|}{Mean Recall}& \multicolumn{2}{|c|}{Accuracy}\\
\hline
& Word2Vec & 71.9 & 79.9 & 70.94 & 53.54\\
& GloVe & 71.9 & 80.3 & 70.11 & 52.18\\
\hline
\end{tabular}
\caption{Preliminary experiments showed GloVe performed similarly to Word2Vec.}
\end{table}

\subsection{Image-Sentence Retrieval Extended Pretrained Embedding Metrics}

\begin{table}[H]
\centering
\small
\setlength{\tabcolsep}{1.5pt}
\begin{tabular}{|rl|c|c|c|c|c|c|c|c|c|c|c|c|c|c|}
\hline
& & \multicolumn{14}{|c|}{Embedding Network~\cite{wangTwoBranch2017}}\\
\cline{3-16}
& & \multicolumn{7}{|c|}{Flickr30K} & \multicolumn{7}{|c|}{MSCOCO}\\
\cline{3-16}
&  & \multicolumn{3}{|c|}{Image-to-Sentence} & \multicolumn{3}{|c|}{Sentence-to-Image} & & \multicolumn{3}{|c|}{Image-to-Sentence} & \multicolumn{3}{|c|}{Sentence-to-Image} &\\
\cline{3-16}
& Method & R@1 & R@5 & R@10 & R@1 & R@5 & R@10 & mR & R@1 & R@5 & R@10 & R@1 & R@5 & R@10 & mR\\
\hline
\hline
{\bf (a)} & \textbf{Training from scratch} & & & & & & & & & & & & & &\\
& Average Embedding & 23.3 & 48.8 & 61.9 & 15.6& 35.3&44.3 &38.2 &55.3 &85.7 &  93.7& 43.7&76.7& 87.1 & 73.7\\
& Self-Attention & 25.9 & 53.4 & 66.2 & 18.1 & 45.5 & 58.8 & 44.6 & 59.8 & 88.7 & 94.9& 45.7 & 79.5 & 90.0 & 76.6\\ 
& LSTM & 45.2 & 72.2 & 82.6 & 29.9 & 59.0&70.9 & 60.0 & 62.8& 89.4& 94.6& 48.1& 81.0 & 89.3 & 77.5\\
\hline
{\bf (b)} & \textbf{Word2Vec} & & & & & & & & & & & & & &\\
& Average Embedding & 47.6 & 75.8 & 84.3 & 31.8 & 62.2 & 73.2 &62.5 & 57.6 & 87.2 & 93.7 & 44.4 & 78.8 & 88.1 & 75.0\\
& Average Embedding + ft & 56.7 & 84.3 & 91.4 & 41.6 & 72.9 & 82.1 & 71.5& 62.4 & 89.1 & 95.0 & 50.2 & 82.2 & 90.2 & 78.2\\
& Self-Attention & 48.7 & 76.0 & 84.5 & 33.0 & 64.4 & 75.2 & 63.6 & 58.6 & 87.4 & 93.2 & 45.4 & 79.7 & 89.4 & 75.6\\ 
& Self-Attention + ft & 57.0 & 84.4 & 91.4 & 42.4 & 73.5 & 82.8 & 71.9 & 64.8 & 91.2 & 96.4 & 51.9 & 83.1 & 91.9 & 79.9\\ 
& LSTM & 50.9 & 81.4&89.3 & 38.9& 70.2 & 80.5 &68.5 & 53.8& 83.4& 92.4& 42.0&76.0  &87.3 &72.5\\
& LSTM + ft & 52.1 & 82.4 & 89.9 & 39.6& 70.0 & 79.9 & 69.0& 63.5 & 89.4& 95.0& 49.7& 81.4 &90.3 &78.2\\
\hline
{\bf (c)} & \textbf{FastText} & & & & & & & & & & & & & &\\
& Average Embedding & 53.3& 82.7 & 90.3 & 39.2 & 70.1	& 80.0 & 69.2 & 62.0 & 91.0 & 96.1& 48.8 & 82.0 & 91.4 & 78.5 \\
& Average Embedding + ft & \textbf{59.4} & \textbf{86.8} & \textbf{92.0} & 42.6 & 73.7 & \textbf{83.5} &73.0 & \textbf{66.6} & 91.7 & 96.6 & 52.7 & \textbf{84.4} & 92.2& \textbf{80.7}  \\
& Self-Attention &53.6 & 81.4 & 90.0 & 40.0 & 71.0 & 81.0 & 69.5 & 63.2 & 90.7 & 95.9 & 48.5 & 82.3 & 91.1 & 78.6\\ 
& Self-Attention + ft & 58.8 & 85.8 & 91.8 & \textbf{44.2} & \textbf{74.6} & 83.3 & \textbf{73.1} & 65.3 & \textbf{92.0} & \textbf{96.7} & \textbf{52.8} & 84.2 & \textbf{92.5} & 80.6\\ 
& LSTM & 52.7&83.3&89.9&38.6&70.2&79.9&69.1 &57.5& 89.7& 95.1 & 47.6 & 81.4&90.6 &76.9\\
& LSTM + ft & 52.1&81.4&89.0&39.0&69.9&79.6& 68.5 & 65.3 &91.5&97.1&51.6&83.7&91.5& 80.1\\
\hline
{\bf (d)} & \textbf{Sentence-Level} & & & & & & & & & & & & & &\\
& InferSent & 56.4 & 54.4 & 91.1 & 40.7 & 72.3 & 82.2 & 71.2 & 60.8 & 90.4 & 96.1 & 47.6 & 77.8 & 85.5 & 76.4\\
& BERT & 57.9 & 84.9 & 91.3 & 41.3 & 73.0 & 82.6 & 71.8 & 58.6 & 89.2 & 95.8 & 46.2 & 76.9 & 85.4 & 75.4\\ 
\hline
\end{tabular}
\caption{Image-sentence retrieval results for pretrained embeddings.}
\end{table}

\subsection{Image-Sentence Retrieval Extended Adapted Embedding Metrics}

\begin{table}[H]
\centering
\small
\setlength{\tabcolsep}{1.5pt}
\begin{tabular}{|rl|c|c|c|c|c|c|c|c|c|c|c|c|c|c|}
\hline
& & \multicolumn{14}{|c|}{Embedding Network~\cite{wangTwoBranch2017}}\\
\cline{3-16}
& & \multicolumn{7}{|c|}{Flickr30K} & \multicolumn{7}{|c|}{MSCOCO}\\
\cline{3-16}
&  & \multicolumn{3}{|c|}{Image-to-Sentence} & \multicolumn{3}{|c|}{Sentence-to-Image} & & \multicolumn{3}{|c|}{Image-to-Sentence} & \multicolumn{3}{|c|}{Sentence-to-Image} &\\
\cline{3-16}
& Method & R@1 & R@5 & R@10 & R@1 & R@5 & R@10 & mR & R@1 & R@5 & R@10 & R@1 & R@5 & R@10 & mR\\
\hline
{\bf (a)} & \textbf{Word2Vec + wn} & & & & & & & & & & & & & &\\
& Average Embedding + ft & 57.7 & 85.3& 91.5& 42.2 & 73.2 & 82.3 & 72.0 & 63.6 & 90.8 & 95.6 & 51.1 & 83.2 & 91.1 & 79.2 \\
& Self-Attention + ft & 57.6 & 86.2 & 92.1 & 42.5 & 73.3 & 82.7 & 72.4 & 64.0 & 91.5 & 96.8 &  51.4 & 84.3 & 91.7 & 80.0\\ 
& LSTM + ft &53.5 &82.8 & 89.9& 39.3& 70.2& 80.5& 69.3& 63.8& 90.6& 95.7&50.2 & 82.0  &90.9 & 78.9\\
\hline
{\bf (b)} & \textbf{GrOVLE} & & & & & & & & & & & & & &\\
& Average Embedding + ft & 57.6 & 85.1 & 92.0 & 42.6 & 73.6 & 82.6 & 72.3 & 65.2 & 91.8 & 96.5 & 52.1 & 83.9 & 92.1 & 80.2 \\
& Self-Attention + ft & 56.9 & 84.2 & 91.7 & 43.2 & 73.9 & 82.8 & 72.1  & \textbf{67.6} & 91.4 & 96.3 & 52.0 & 83.7 & 92.1 & 80.5\\ 
& LSTM + ft & 54.1 & 82.7 & 91.1 & 39.7 & 70.2& 80.1 & 69.7& 65.0 & 89.6 & 95.8 & 49.7 & 82.0 & 90.8 & 78.8\\
\hline
{\bf (c)} & \textbf{Visual Word2Vec} & & & & & & & & & & & & & &\\
& Average Embedding + ft & 50.0 &79.7&87.0&37.0&68.3&78.6& 66.8 &61.7&90.6&95.8&50.0&82.7&91.2& 78.7\\
& Self-Attention + ft & 51.3 & 82.3 & 89.5 & 40.9 & 69.1 & 79.9 & 68.8  & 61.6 & 91.4 & 96.7 & 50.2 & 83.1 & 92.4 & 79.2\\ 
& LSTM + ft & 50.5 & 78.3 & 88.6& 36.2 &67.7 & 78.7 & 66.7 & 56.2&87.3&94.8&42.5&77.3&87.8&74.5 \\
\hline
{\bf (d)} & \textbf{HGLMM (300-D)} & & & & & & & & & & & & & &\\
& Average Embedding + ft & 56.6 & 84.2 & 90.8& 41.4& 72.0& 81.2& 71.0&65.5 & 90.7& 96.0 & 51.5&  83.4&91.5 &79.8 \\
& Self-Attention + ft & 56.4 & 84.7  &  91.3 & 42.1 & 73.3  & 82.2 & 71.8 & 66.2 & 91.0 &  96.3& 51.8 & \textbf{84.7} & \textbf{92.6} & 80.4 \\ 
& LSTM + ft & 54.1 & 82.0 & 90.2 & 40.2 & 70.4& 80.2 & 69.5& 61.5 & 89.9 & 95.3& 48.9&  81.5& 90.4&77.9\\
\hline
{\bf (e)} & \textbf{HGLMM (6K-D)} & & & & & & & & & & & & & &\\
& Average Embedding + ft & 60.5& 86.4& 92.9& 43.8&73.9 &83.3 &73.5 &67.2 &91.7 &\textbf{97.5} & \textbf{53.0}  & 84.0& 92.2 & \textbf{80.9} \\
& Self-Attention + ft & \textbf{61.6} & \textbf{88.4} & \textbf{94.5}& \textbf{46.4} & \textbf{75.7} & \textbf{84.1} & \textbf{75.1} & 65.4 & \textbf{93.0} & 97.4 & 52.6 & 83.6 & 90.6 & 80.6\\ 
& LSTM + ft & 51.4 & 80.7 & 89.4 & 39.1 & 68.7 & 78.6 & 68.0 & 65.0 & 90.7 & 96.1& 51.2& 82.8 & 90.9 & 79.4\\
\hline
\end{tabular}
\caption{Image-sentence retrieval results for adapted embeddings.}
\end{table}
\subsection{Image-Sentence Retrieval Extended Multi-task Trained GrOVLE Metrics}

\begin{table}[H]
\centering
\small
\setlength{\tabcolsep}{1.5pt}
\begin{tabular}{|rl|c|c|c|c|c|c|c|c|c|c|c|c|c|c|}
\hline
& & \multicolumn{14}{|c|}{Embedding Network~\cite{wangTwoBranch2017}}\\
\cline{3-16}
& & \multicolumn{7}{|c|}{Flickr30K} & \multicolumn{7}{|c|}{MSCOCO}\\
\cline{3-16}
&  & \multicolumn{3}{|c|}{Image-to-Sentence} & \multicolumn{3}{|c|}{Sentence-to-Image} & & \multicolumn{3}{|c|}{Image-to-Sentence} & \multicolumn{3}{|c|}{Sentence-to-Image} &\\
\cline{3-16}
& Method & R@1 & R@5 & R@10 & R@1 & R@5 & R@10 & mR & R@1 & R@5 & R@10 & R@1 & R@5 & R@10 & mR\\
\hline
\hline
& GrOVLE w/o multi-task pretraining & 47.3  &  78.9 &  87.0& 33.2& 65.1 & 76.8 & 64.7 & 56.3 & 87.4 & 94.3 & 44.5 & 79.0 & 88.5 & 75.0\\
& + multi-task pretraining w/o target task & 49.0 & 79.7 & 87.7 & 35.7 & 66.2 & 76.3 & 65.8 & 60.8 & 87.3 & 94.7 & 46.7 & 79.7 & 89.3 & 76.4\\
& + multi-task pretraining w/ target task & 51.3 & 68.7 & 80.7 & 36.2 & 64.3 & 66.3 & 66.2 & 65.5 & 91.6 & 96.7 & 51.2 & 83.6 & 91.4 & 80.2\\ 
& + multi-task pretraining w/ target task + ft& \textbf{58.2} & \textbf{85.8} & \textbf{91.9} & \textbf{42.1} & \textbf{73.8} & \textbf{84.0} & \textbf{72.6} & \textbf{66.8} & \textbf{93.4} & \textbf{97.9}&  \textbf{51.8}& \textbf{85.0}& \textbf{92.8}& \textbf{81.3}\\
\hline
\end{tabular}
\caption{Image-sentence retrieval results for multi-task trained GrOVLE, created using the original set of task models.}
\end{table}

\subsection{Image-Sentence Retrieval Additional Model Metrics}

\begin{table}[H]
\centering
\small
\setlength{\tabcolsep}{1.5pt}
\begin{tabular}{|rl|c|c|c|c|c|c|c|c|c|c|c|c|c|c|}
\hline
& & \multicolumn{14}{|c|}{Stacked Cross Attention Network (SCAN) \cite{lee2018stacked}} \\
\cline{3-16}
& & \multicolumn{7}{|c|}{Flickr30K} & \multicolumn{7}{|c|}{MSCOCO}\\
\cline{3-16}
&  & \multicolumn{3}{|c|}{Image-to-Sentence} & \multicolumn{3}{|c|}{Sentence-to-Image} & & \multicolumn{3}{|c|}{Image-to-Sentence} & \multicolumn{3}{|c|}{Sentence-to-Image} &\\
\cline{3-16}
& Method & R@1 & R@5 & R@10 & R@1 & R@5 & R@10 & mR & R@1 & R@5 & R@10 & R@1 & R@5 & R@10 & mR\\
\hline
\hline
& Training from scratch & 60.8 & 86.8 & 92.0 & 43.0 & 72.1 & 81.9 & 72.8 & 69.9 & 94.3 & 97.4 & 56.6 & 87.1 & 94.0 & 83.2\\
& Word2Vec + ft & 59.7 & 83.4 & 90.9 & 41.2 & 70.6 & 79.8 & 70.9 & 71.9 & 94.1 & \textbf{98.1} & 58.2 & \textbf{87.8} & 93.8 & 84.0\\
& FastText + ft & 60.7 & 86.8 & 91.5 & 42.1 & 73.0 & 80.8 & 72.5 & 71.4 & 94.4 & 97.7 & 58.0 & 87.4 & 93.8 & 83.8 \\
& GrOVLE (w/o multi-task pretraining) + ft & 61.0 & 86.7 & 92.0 & 42.2 & 72.7 & 81.3 & 72.7 & 72.3 & 94.0 & 97.9 & 58.4 & 87.7 & \textbf{94.4} & 84.1\\
& + multi-task pretraining w/ target task + ft & \textbf{65.8} & \textbf{89.8} & \textbf{94.2} & \textbf{46.8} & \textbf{76.2} & \textbf{84.5} & \textbf{76.2} & \textbf{74.4} & \textbf{94.8} & 97.8 & \textbf{59.1} & \textbf{87.8} & 94.2 & \textbf{84.7}\\
\hline
\end{tabular}
\caption{Image-sentence retrieval results with the additional retrieval model for from-stratch, Word2Vec, FastText, GrOVLE, and multi-task trained GrOVLE representations. The multi-task trained GrOVLE was created from the full set of additional models.}
\end{table}
\subsection{Phrase Grounding Additional Model Metrics}

\begin{table}[H]
    \small
    \setlength{\tabcolsep}{1.4pt}
    \centering
    \begin{tabular}{|rl|c|c|}
     \hline
     & & \multicolumn{2}{|c|}{Query Adaptive R-CNN \cite{hinamiEMNLP2018}} \\
     \cline{3-4}
    & & Flickr30k Entities & ReferIt\\
    \cline{3-4}
    & Method & \multicolumn{2}{|c|}{Accuracy}\\
    \hline
    & Training from scratch & 68.56 & 50.23\\
    & Word2Vec + ft & 69.78 &  52.97\\
    & FastText + ft & 69.27  & 53.01\\
    & BERT & 66.30 &  51.09\\
    & GrOVLE (w/o multi-task pretraining) + ft & 70.03 &  53.88\\
    & + multi-task pretraining w/ target task + ft & \textbf{71.08}& \textbf{54.10}\\
\hline
    \end{tabular}
    \caption{Phrase grounding results with the additional grounding model for from-stratch, Word2Vec, FastText, BERT, GrOVLE, and multi-task trained GrOVLE representations. The multi-task trained GrOVLE was created from the full set of additional models.}
    \label{tab:imagecap_multi}
\end{table}

\subsection{Text-to-Clip Extended Pretrained Embedding Metrics}

\begin{table}[H]
    \small
    \setlength{\tabcolsep}{2pt}
    \centering
    \begin{tabular}{|rl|c|c|c|c|}
    \hline
    & & \multicolumn{4}{|c|}{CITE~\cite{plummerCITE2017}}\\
    \cline{3-6}
    & Method & R@1 & R@5 & mIOU & Average\\
    \hline
    {\bf (a)} & \textbf{Training from scratch} & & & & \\
& Average Embedding & 15.53 & 58.21 & 25.32 & 33.02\\
& Self-Attention & 15.41 & 57.85 & 27.17 & 33.48\\ 
& LSTM & 14.38 & 59.02 & 25.08 & 32.83\\
\hline
{\bf (b)} & \textbf{Word2Vec} & & & & \\
& Average Embedding & 15.91 & 56.08 & 26.85 & 32.95\\
& Average Embedding + ft & 15.65 & 55.00 & 27.10 & 32.58\\
& Self-Attention & 15.87 & 55.89 & 27.90 & 33.23\\ 
& Self-Attention + ft & 15.81 & 55.48 & \textbf{28.48} & 33.26\\ 
& LSTM & \textbf{16.27} & 57.94 & 26.97 & 33.73\\
& LSTM + ft & 15.49 & 59.29 & 25.04 & \textbf{33.94}\\
\hline
{\bf (c)} & \textbf{FastText} & & & &\\
& Average Embedding & 15.22 & 56.08 & 26.06 & 32.45 \\
& Average Embedding + ft & 15.69 & 53.72 & 26.62 & 32.01\\
& Self-Attention & 15.92 & 56.14 & 27.87 & 33.31\\ 
& Self-Attention + ft & 15.60 & 55.93 & 27.99 & 33.17\\ 
& LSTM & 14.40 & \textbf{60.21} & 24.56 & 33.06\\
& LSTM + ft & 14.80 & 58.02& 24.71& 32.51\\
\hline
{\bf (d)} & \textbf{Sentence-Level} & & & &\\
& InferSent & 14.33 & 56.10 & 25.18 & 31.87\\
& BERT & 14.23 & 58.76 & 24.39 & 32.46\\
\hline
    \end{tabular}
    \caption{Text-to-clip results for pretrained embeddings on DiDeMo.}
    \label{tab:ttc_pretrain}
\end{table}

\subsection{Text-to-Clip Extended Adapted Embedding Metrics}

\begin{table}[H]
    \small
    \setlength{\tabcolsep}{2pt}
    \centering
    \begin{tabular}{|rl|c|c|c|c|}
    \hline
    & & \multicolumn{4}{|c|}{CITE~\cite{plummerCITE2017}}\\
    \cline{3-6}
    & Method & R@1 & R@5 & mIOU & Average\\
    \hline
{\bf (a)} & \textbf{Word2Vec + wn} & & & & \\
& Average Embedding + ft & 16.05 & 55.89 & 27.79 & 33.24\\
& Self-Attention + ft & 16.05 & 57.73 & 27.16 & 33.65\\ 
& LSTM + ft & 16.36 & 59.81 & 26.32 & 34.16\\
\hline
{\bf (b)} & \textbf{GrOVLE} & & & & \\
& Average Embedding + ft & \textbf{16.53} & 56.05 & \textbf{28.56} & 33.71\\
& Self-Attention + ft & 15.60 & 58.16 & 25.67 & 33.14\\ 
& LSTM + ft & 15.79 & \textbf{61.65} & 25.98 & 34.47\\
\hline
{\bf (c)} & \textbf{Visual Word2Vec} & & & &\\
& Average Embedding + ft &14.05 &56.90 & 24.23 &31.73\\
& Self-Attention + ft & 14.12 & 55.23 & 24.11 &31.15\\ 
& LSTM + ft & 14.03 & 58.52 &24.31  &32.29\\
\hline
{\bf (d)} & \textbf{HGLMM (300-D)} & & & &\\
& Average Embedding + ft & 15.96 & 54.67 & 27.24 & 32.62\\
& Self-Attention + ft & 16.23 & 56.07 & 28.01 & 33.44\\ 
& LSTM + ft & 15.89 & 59.84 & 25.81 & 33.85\\
\hline
{\bf (e)} & \textbf{HGLMM (6K-D)} & & & &\\
& Average Embedding + ft & 15.43 & 55.79 & 26.76 & 32.66\\
& Self-Attention + ft & 15.60 & 57.82 & 27.30 & 33.57\\
& LSTM + ft & 16.41 & 60.86 & 26.59 & \textbf{34.62}\\
\hline
    \end{tabular}
    \caption{Text-to-clip results for adapted embeddings on DiDeMo.}
    \label{tab:ttc_pretrain}
\end{table}

\subsection{Text-to-Clip Extended Multi-task Trained GrOVLE Metrics}

\begin{table}[H]
    \small
    \setlength{\tabcolsep}{2pt}
    \centering
    \begin{tabular}{|rl|c|c|c|c|}
    \hline
    & & \multicolumn{4}{|c|}{CITE~\cite{plummerCITE2017}}\\
    \cline{3-6}
    & Method & R@1 & R@5 & mIOU & Average\\
    \hline
    & GrOVLE w/o multi-task pretraining & 16.34 & \textbf{60.84} & 26.17 & 34.45\\
& + multi-task pretraining w/o target task & 16.94 & 58.90 & 27.88 & 34.57\\
& + multi-task pretraining w/ target task & 16.96 & 59.40 & 28.09 & 34.82\\ 
& + multi-task pretraining w/ target task + ft& \textbf{17.05} & 59.84 & \textbf{28.39}& \textbf{35.09}\\
    \hline
    \end{tabular}
    \caption{Text-to-clip results for multi-task trained GrOVLE on DiDeMo.}
    \label{tab:ttc_pretrain}
\end{table}

\subsection{Text-to-Clip Additional Model Metrics}

\begin{table}[H]
    \small
    \setlength{\tabcolsep}{1.4pt}
    \centering
    \begin{tabular}{|rl|c|c|c|c|}
     \hline
     & & \multicolumn{4}{|c|}{Temporal GroundNet (TGN) \cite{chenEMNLP2018}} \\
    \cline{3-6}
    & Method & R@1 & R@5 & mIOU & Average\\
    \hline
    & Training from scratch & \textbf{26.26} & \textbf{74.33} & 31.32 & 43.97\\
    & Word2Vec + ft & 25.98 & 74.11 & 32.06 & 44.05 \\
    & FastText + ft & 26.13 & 74.23 & 30.53 & 43.64 \\
& GrOVLE (w/o multi-task pretraining) + ft& 25.54 & 73.98 & \textbf{34.24} & \textbf{44.59} \\
& + multi-task pretraining w/ target task + ft  & 24.91 & 73.58 & 32.37 & 43.62\\
\hline
    \end{tabular}
    \caption{Text-to-clip results with the additional text-to-clip model for from-stratch, Word2Vec, FastText, GrOVLE, and multi-task trained GrOVLE representations on DiDeMo. The multi-task trained GrOVLE was created from the full set of additional models.}
    \label{tab:imagecap_multi}
\end{table}
\subsection{Image Captioning Extended Pretrained Embedding Metrics}

\begin{table}[H]
    \small
    \setlength{\tabcolsep}{1.4pt}
    \centering
    \begin{tabular}{|rl|c|c|c|}
    \hline
    & & \multicolumn{3}{|c|}{ARNet~\cite{tencent}} \\
    \cline{3-5}
    & Method & BLEU-4  & CIDER & METEOR\\
    \hline
    \textbf{(a)} & \textbf{Training from scratch} & &&\\
    & LSTM &  -- & -- & --  \\
    & LSTM + ft & 26.7&89.7&24.3\\
        \hline
 \textbf{(b)} & \textbf{Word2Vec}  & & & \\
 & LSTM &28.1&92.7&24.7 \\
  & LSTM + ft &\textbf{28.5}&\textbf{94.0}&\textbf{24.8}\\
      \hline
    \textbf{(c)} & \textbf{FastText} & & &\\
    & LSTM & \textbf{28.5}&92.7&24.7 \\
  & LSTM + ft & 28.3&93.2&\textbf{24.8}\\
    \hline
    \end{tabular}
    \caption{Image captioning results for pretrained embeddings on MSCOCO.}
    \label{tab:imagecap_pretrained}
\end{table}

\subsection{Image Captioning Extended Adapted Embedding Metrics}

\begin{table}[H]
    \small
    \setlength{\tabcolsep}{1.4pt}
    \centering
    \begin{tabular}{|rl|c|c|c|}
    \hline
    & & \multicolumn{3}{|c|}{ARNet~\cite{tencent}} \\
    \cline{3-5}
    & Method & BLEU-4  & CIDER & METEOR\\
    \hline
    \textbf{(a)} & \textbf{Word2Vec + wn} & & &\\
    & LSTM + ft &28.6&93.3&\textbf{24.9}\\
        \hline
 \textbf{(b)} & \textbf{GrOVLE}  & & & \\
  & LSTM + ft & 28.3&92.5&24.8\\
      \hline
    \textbf{(c)} & \textbf{Visual Word2Vec} & & &\\
  & LSTM + ft & \textbf{28.8}&\textbf{94.0}&\textbf{24.9}\\
    \hline
        \textbf{(c)} & \textbf{HGLMM (300-D)} & & &\\
  & LSTM + ft &28.7&\textbf{94.0}&\textbf{24.9}\\
      \hline
        \textbf{(c)} & \textbf{HGLMM (6K-D)} & & & \\
  & LSTM + ft & 28.0&92.8&24.7\\
  \hline
    \end{tabular}
    \caption{Image captioning results for adapted embeddings on MSCOCO.}
    \label{tab:imagecap_adapted}
\end{table}

\subsection{Image Captioning Extended Multi-task Trained GrOVLE Metrics}

\begin{table}[H]
    \small
    \setlength{\tabcolsep}{1.4pt}
    \centering
    \begin{tabular}{|rl|c|c|c|}
    \hline
    & & \multicolumn{3}{|c|}{ARNet~\cite{tencent}} \\
    \cline{3-5}
    & Method & BLEU-4  & CIDER & METEOR\\
    \hline
& GrOVLE w/o multi-task pretraining & 28.5&92.7&\textbf{24.7}\\
& + multi-task pretraining w/o target task & \textbf{28.8}&\textbf{93.3}&\textbf{24.7} \\
& + multi-task pretraining w/ target task& 28.5& 92.7& \textbf{24.7}\\ 
& + multi-task pretraining w/ target task + ft&28.7&93.2&\textbf{24.7} \\
\hline
    \end{tabular}
    \caption{Image captioning results for multi-task trained GrOVLE on MSCOCO.}
    \label{tab:imagecap_multi}
\end{table}

\subsection{Image Captioning Additional Model Metrics}
\begin{table}[H]
    \small
    \setlength{\tabcolsep}{1.4pt}
    \centering
    \begin{tabular}{|rl|c|c|c|}
    \hline
    & & \multicolumn{3}{|c|}{Neural Image Captioning (NIC) \cite{vinyalsCVPR2015}}\\
    \cline{3-5}
    & Method & BLEU-4  & CIDER & METEOR\\
    \hline
        & Training from scratch & 18.2 & 62.5 & 20.3\\
    & Word2Vec + ft & 18.7 & 62.8 & 20.2 \\
    & FastText + ft & 17.9 & 61.6& 17.9\\
& GrOVLE (w/o multi-task pretraining) + ft & \textbf{19.4} & \textbf{65.4} & 20.6 \\
& + multi-task pretraining w/ target task + ft & \textbf{19.4} & 65.1 & \textbf{20.9} \\
\hline
    \end{tabular}
    \caption{Image captioning results with an additional captioning model for from-stratch, Word2Vec, FastText, GrOVLE, and multi-task trained GrOVLE representations on MSCOCO. The multi-task trained GrOVLE was created from the full set of additional models.}
    \label{tab:imagecap_multi}
\end{table}

\begin{table}[H]
    \small
    \setlength{\tabcolsep}{1.4pt}
    \centering
    \begin{tabular}{|rl|c|c|c|}
     \hline
     & & \multicolumn{3}{|c|}{Bottom-Up Top-Down Attention (BUTD) \cite{andersonCVPR2018}}\\
    \cline{3-5}
    & Method & BLEU-4  & CIDER & METEOR\\
    \hline
    & Training from scratch & 35.2& 109.8 & 27.2\\
    & Word2Vec + ft& 35.1 & 110.8& 27.1\\
    & FastText + ft & 35.2 & 110.3& 27.1\\
& GrOVLE (w/o multi-task pretraining) + ft & 35.1 & 110.4 & 27.1\\
& + multi-task pretraining w/ target task + ft  &\textbf{35.7} & \textbf{111.6} & \textbf{27.3}\\
\hline
    \end{tabular}
    \caption{Image captioning results with an additional captioning model for from-stratch, Word2Vec, FastText, GrOVLE, and multi-task trained GrOVLE representations on MSCOCO. The multi-task trained GrOVLE was created from the full set of additional models.}
    \label{tab:imagecap_multi}
\end{table}

\subsection{Visual Question Answering Additional Model Metrics}

\begin{table}[H]
    \small
    \setlength{\tabcolsep}{1.4pt}
    \centering
    \begin{tabular}{|rl|c|c|}
     \hline
     & & \multicolumn{2}{|c|}{Bilinear Attention Network} \\
     & & \multicolumn{2}{|c|}{(BAN) \cite{kim2018bilinear}} \\
    \cline{3-4}
    & Method & \multicolumn{2}{|c|}{Accuracy} \\
    \hline
    & Training from scratch & \multicolumn{2}{|c|}{68.68} \\
    & Word2Vec + ft & \multicolumn{2}{|c|}{69.91} \\
    & FastText + ft & \multicolumn{2}{|c|}{69.91} \\
& GrOVLE (w/o multi-task pretraining) + ft & \multicolumn{2}{|c|}{69.36} \\
& + multi-task pretraining w/ target task + ft & \multicolumn{2}{|c|}{\textbf{69.97}} \\
\hline
    \end{tabular}
    \caption{Visual Question Answering results with the additional VQA model for from-stratch, Word2Vec, FastText, GrOVLE, and multi-task trained GrOVLE representations on VQA v2. The multi-task trained GrOVLE was created from the full set of additional models.}
    \label{tab:imagecap_multi}
\end{table}
\clearpage

\end{document}